\documentclass{article}

\usepackage{microtype}
\usepackage{graphicx}
\usepackage{subfigure}
\usepackage{booktabs} 

\usepackage{hyperref}

\usepackage{subfigure}

\usepackage{mathtools}
\usepackage{amssymb}

\usepackage{shortcuts}
\usepackage{amsmath}

\usepackage[accepted]{icml2021}

\icmltitlerunning{Inductive biases, pretraining and fine-tuning jointly account for brain responses to speech}

\begin{document}

\twocolumn[
\icmltitle{Inductive biases, pretraining and fine-tuning jointly account for brain responses to speech}




\begin{icmlauthorlist}
\icmlauthor{Juliette MILLET}{fb,llf,coml,cri}
\icmlauthor{Jean-Rémi KING}{fb,psl}
\end{icmlauthorlist}

\icmlaffiliation{fb}{Facebook AI Research, Paris}
\icmlaffiliation{psl}{École normale supérieure, PSL University, CNRS, Paris, France}
\icmlaffiliation{llf}{Universit{\'{e}} de Paris, LLF, CNRS, Paris, France}
\icmlaffiliation{coml}{CoML, ENS/CNRS/EHESS/INRIA/PSL University, Paris, France}
\icmlaffiliation{cri}{CRI, Département Frontières du Vivant et de l'Apprendre, IIFR, Universit{\'{e}} de Paris}

\icmlcorrespondingauthor{Juliette MILLET}{juliette.millet@cri-paris.org}

\icmlkeywords{keywords}

\vskip 0.3in
]



\printAffiliationsAndNotice{}

\begin{abstract}

Our ability to comprehend speech remains, to date, unrivaled by deep learning models. This feat could result from the brain’s ability to fine-tune generic sound representations for speech-specific processes. To test this hypothesis, we compare i) five types of deep neural networks to ii) human brain responses elicited by spoken sentences and recorded in 102 Dutch subjects using functional Magnetic Resonance Imaging (fMRI). Each network was either trained on an acoustics scene classification, a speech-to-text task (based on Bengali, English, or Dutch), or not trained. The similarity between each model and the brain is assessed by correlating their respective activations after an optimal linear projection. The differences in brain-similarity across networks revealed three main results. First, speech representations in the brain can be accounted for by random deep networks. Second, learning to classify acoustic scenes leads deep nets to increase their brain similarity. Third, learning to process phonetically-related speech inputs (i.e., Dutch vs English) leads deep nets to reach higher levels of brain-similarity than learning to process phonetically-distant speech inputs (i.e. Dutch vs Bengali). Together, these results suggest that the human brain fine-tunes its heavily-trained auditory hierarchy to learn to process speech.

\end{abstract}

\section{Introduction}

To understand speech, the human brain continuously transforms and parses the stream of auditory features generated by the cochleas. To date, this ability remains unrivaled by deep learning architectures, especially in noisy or accented speech conditions \cite{zhang2020learning, unni2020coupled}, partly because the exact nature of speech representations remains underdetermined. Indeed, while the brain -- and the superior temporal cortex in particular -- represents and clusters phonetic features \citep{mesgarani2014phonetic,kell2018task}, the corresponding representations may not be specific to speech but generic to sounds \cite{galantucci2006motor,daube2019simple}. It is thus unclear whether speech models should be trained with speech input only or from a wide variety of sounds.

\begin{figure*}[h!]
    \centering
    \vskip 0.2in
    \includegraphics[height=2.51in]{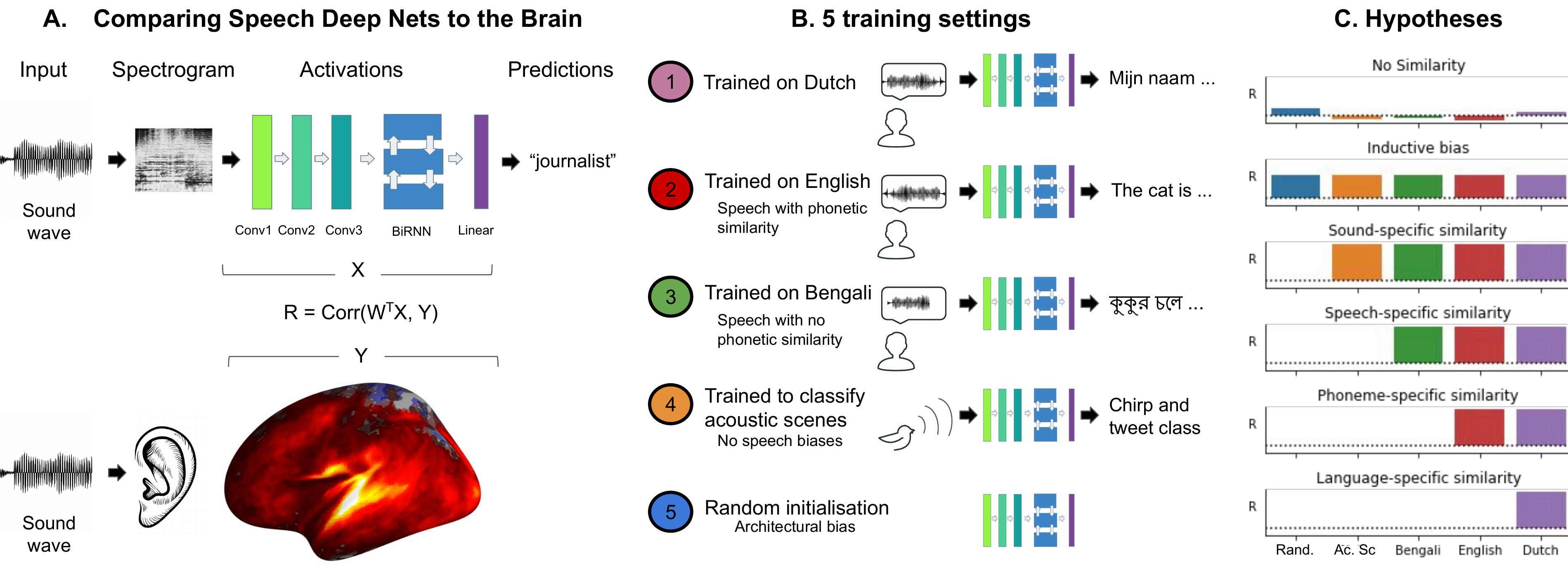}
    \caption{\textbf{A.} The functional similarity between an artificial neural network and the human brain was quantified by correlating i) brain responses to speech to ii) the networks'activations in response to the same input, and optimally projected with a linear operator $W$ onto the subject brain. To this end, a linear ridge regression ($W$) is fit from the model’s activations convolved by an HRF ($X$) to the brain response ($Y$) to the same stimulus sequence. The resulting "brain score" \cite{yamins2014performance} is a Pearson correlation between true brain activity and the brain activity predicted by a linear projection of the model's activation. \textbf{B.} Four types of networks, using the same architecture were trained on Dutch, English, and Bengali speech-to-text and on a acoustic scene classification, respectively. The fifth network corresponds to an untrained version, randomly initialised. Each type was trained with six random seeds, leading to a total of 30 networks. \textbf{C.} The neural networks will be considered to show no similarity with brain activity if they show brain scores (R) not significantly above zero. }
    \label{fig:explanation}
    \vskip -0.2in
\end{figure*}

We propose to address this issue by comparing i) the brain representations elicited during speech comprehension to ii) the representations of artificial neural networks, input with the same sounds, and trained in three distinct regimes: random neural networks (i.e. not trained), neural networks trained to process generic sounds (i.e. acoustic scene classification), and neural networks trained to process speech  (i.e. speech-to-text using Dutch, English or Bengali). Previous studies have already shown that deep convolutional neural networks trained to classify words, musical genres \cite{kell2018task, kumar2020searching} or natural sounds \cite{koumura2019cascaded}, generate brain-like representations, in the sense that one can find a linear correspondence between the activation of the neural networks and the activations of the brain (Figure \ref{fig:explanation}). This similarity can be quantified with a "brain score" \cite{yamins2014performance}, a correlation between the brain measurements and a linear projection of the model's activations, under the assumption that representations are defined as linearly exploitable information \citep{hung2005fast,kamitani2005decoding,kriegeskorte2013representational,king2018encoding}. We thus hypothesize that the nature of speech representations in the brain can be elucidated by comparing them to those of random, sound-generic and speech-specific neural networks (Figure \ref{fig:explanation}).  


To this end, we analyze the functional Magnetic Resonance Imaging (fMRI) recordings of 102 Dutch speakers who each underwent a 1\,h speech comprehension task, consisting of sequences of unrelated sentences. We then compared, for each subject separately, the resulting blood oxygenation level dependent (BOLD) responses to the activations of 30 deep neural networks, all derived from the same architecture taken from DeepSpeech 2  \cite{amodei2016deep}.

We formalize our hypotheses with the following non-mutually exclusive predictions (Figure \ref{fig:explanation}): the brain representations of speech are likely to benefit from an \textbf{inductive bias} if a random neural network can generate representations similar to those of the brain, \textbf{generic} if a neural network trained on non-speech sounds leads to representations that are more similar to brain activity than a random network, \textbf{speech-specific} if a neural network trained on speech sounds leads to representations that are more similar to brain activity than a network trained on non-speech sounds, \textbf{phoneme-specific} if a neural network trained on a language that roughly contains the same speech sounds (or phonemes) than the participants' native language (e.g. English vs Dutch) leads to representations that are more similar to brain activity than a network trained on a language that has a very different phonemes inventory (e.g. Bengali vs Dutch), and \textbf{language-specific} if a neural network trained on the same language as the participants' native language (i.e. Dutch) leads to representations that are more similar to brain activity than a network trained on a distinct, but phonetically-related language (e.g. Dutch vs English).



%
Our results support several of these predictions. First, the architecture of the network, without training, suffices to significantly account for the hierarchy of brain responses to speech. Second, training on auditory-scene classification increases the brain-similarity of the networks. Third, training on English leads to higher brain scores than training on Bengali. Together, these results suggest that the brain combines inductive biases, generic, and phoneme-specific representations to process speech. 

\section{Methods}
\label{methods}



\subsection{Stimuli}
One-hundred and two subjects performed a simple speech comprehension task while being recorded with functional Magnetic Resonance Imaging (fMRI) by Schoffelen and colleagues \cite{schoffelen2019204}. Each participant listened to 60 isolated sentences, grouped in series of five and selected from a pool of 360 sentences, varying from 9 to 15 words. Each sentence lasted on average $4.28\pm0.6$ seconds (maximal length: 6s, minimal length: 2.8s), for a total of 28 min and 49 seconds. All sentences were recorded by a unique female native Dutch speaker in stereo at 44.1\,kHz. We downsample the recordings to 16\,kHz and convert them to mono with the Sox software \footnote{\url{http://sox.sourceforge.net/}}. Note that the dataset also contains word lists that are not presently analyzed. However, these sound sequences included silences between each word and thus lead to abnormal speech inputs.

\subsection{fMRI preprocessing}
Data were provided (in part) by the Donders Institute for Brain, Cognition and Behaviour.
Each subject performed two MRI scans performed: a structural T1-weighted magnetization-prepared rapid gradient-echo (MP-RAGE) pulse sequence (TR=2,300 ms, TE=3.03 ms, 8 degree flip-angle, 1 slab, slice-matrix size=256×256, slice thickness=1 mm, field of view=256 mm, isotropic voxel-size=1.0×1.0×1.0 mm) and a functional T2$^*$-weighted echo planar blood oxygenation level dependent (EPI-BOLD) sequence (TR=2.0 seconds, TE=35ms, flip angle=90 degrees, anisotropic voxel size=3.5×3.5×3.0 mm extracted from 29 oblique slices.) Additional acquisition details can be found in \citet{schoffelen2019204}.

Structural images were defaced by Schoffelen and colleagues. We then preprocessed them with Freesurfer \citep{fischl2012freesurfer}, using the \texttt{recon-all} pipeline with default parameters and a manual inspection of the cortical segmentations. Region-of-interest analyses of regions the primary auditory cortex A1, Belt (LBelt MBelt PBelt) and TPJ (STV, PFm, PGi) were selected from the PALS Brodmann' area atlas \citep{van2005population} and the Destrieux Atlas segmentation \citep{destrieux2010automatic}. 

Functional images are preprocessed with fMRIPrep with default parameters \citep{fmriprep}. 
The Blood Oxygenation Level Dependent (BOLD) time series are detrended and de-confounded from 18 variables: the six head motion parameters (\texttt{trans$_{x,y,z}$}, \texttt{rot$_{x,y,z}$}) estimated analytically, the first 6 noise components derived from CompCorr \citep{behzadi2007component} and the 6 DCT-basis regressors derived from nilearn's \texttt{clean\_img} pipeline and otherwise default parameters \citep{nilearn}.
The resulting volumetric data are projected onto the surface, after subselecting voxels along a 3mm ``line" orthogonal to the mid-thickness. Surface projections are spatially decimated by 10, and are hereafter referred to as voxels, for simplicity. 

To mitigate slow drifts in fMRI acquisition, the recordings of each group of 5 sentences are separately and linearly detrended. The 12 cross-validation splits are applied across 5-sentences blocks (i.e., across detrended segments) to ensure no information leakage between the training and testing sets.


%


\subsection{Networks}
We train and analyze the activations of five distinct types of neural networks (training details are in the next section). All networks share the same architecture derived from Deepspeech 2 \cite{amodei2016deep}, and are all input with spectrograms $U \in \bbR^{d_\tau \times d_u}$ ($d_u=160$) of mono-sound waveforms computed with librosa's \texttt{stft} function \cite{librosa} squared using: window size 20 milliseconds, window stride of 10 ms, and \texttt{n\_ftt} = 320. These spectrograms are transformed by a hierarchy of three 2D convolutional layers (with Layer 1: kernel $k_1=11\times41$, stride $s_1=2\times2$, number of channels $c_1=32$; Layer 2: $k_2=11\times21$, $s_2=1\times2$, $c_2=64$, Layer 3: $k_3=11\times21$, $s_3=1\times2$, $c_3=96$), followed by one bidirectional gated recurrent unit (GRU, 1 layer with dimension 256) and ending in a fully connected layer, whose dimensionality varies with the task (see below). A rectified linear unit (ReLU) function is used between each layer. Finally, a softmax layer is applied on the linear layer' outputs to obtain a probability distribution over characters or acoustic-scene classes. 

\subsection{Tasks}
\label{sec:task}
To test our hypotheses, We train each network on one of four different datasets (Figure \ref{fig:explanation}). The Dutch and the English datasets are taken from the aligned part of the Spoken Wikipedia Corpora \cite{Baumann2018}. The Bengali dataset is taken from \citet{kjartansson-etal-sltu2018}. In order to have approximately the same number of characters possible for the three languages we romanize the Bengali transcriptions using URoman \cite{hermjakob-etal-2018-box}. Finally, the acoustic scene classification dataset is extracted from the curated subset of the FSDKaggle2019 dataset \cite{fonseca2019audio} with manual annotations \cite{fonseca2017freesound}. 

We use 66 hours and 20 minutes (roughly 100,000 sentences) of the first three datasets to match the Dutch dataset, and ensure that the networks had a similar data exposure. We restrict the scene acoustic dataset to its curated version (10 hours long) to avoid introducing noisy labels. Each dataset is then divided into a training (80\%), a validation (10\%) and a test set (10\%). Input samples are limited to a maximum length of 20 seconds.

Each model $m$ was trained using Connectionist Temporal Classification (CTC) \cite{graves2006connectionist} parameterized over $\theta$:

\begin{equation}
    \argmin_{\theta} -\log \sum_{a \in a_{U,V}} \prod_{t}^{d_t} p_{CTC}\left(a_{t} \mid m_\theta(U)\right)
\end{equation}

where $m_\theta(U) \in \bbR^{d_\tau \times d_v}$ are the probabilistic predictions of the networks at each $\tau$ time sample given the spectrograms $U\in \bbR^{d_\tau \times d_u}$, $V \in \bbR^{d_t \times d_v}$ are the true transcriptions of $U$, and $a_{U,V}$ is the set of all possible alignments between $U$ and $V$.

For the models trained on speech recognition, $V$ is a textual transcription. Consequently, $d_v$ is the number of possible characters: $d_v=37$ for Dutch, $d_v=28$ for English and Bengali. For the models trained on acoustic scene classification, $V$ is a one hot vector indicating the class of $U$: i.e. $d_v=80$ classes.

The models are trained with Openseq2seq \cite{openseq2seq}, with six different initialisation seeds for each dataset, using stochastic gradient descent with the following optimization hyperparameters: momentum of $0.9$, an initial learning rate of $0.001$, a polynomial decay policy with power $2$, a batch size of $8$. After each gradient descent, we select the model based on its performance on the validation set, with a maximum number of training steps of 70,000. For each setting, we compute the brain scores for each initialisation and present the average results across initialisations, for simplicity.

On average, the Dutch, English and Bengali models achieved a Word Error Rate (WER) of 0.59, 0.78, and 0.85, on their respective test sets. The acoustic scene classification models achieve a class error rate of 0.64. These modest performances results from a design choice: to avoid semantic representations, we deliberately decided not to incorporate a pretrained language model feeding the architecture. 

\subsection{Preprocessing of model activations}
The activations of the models are sampled at a different frequency ($1/t=50$\,Hz) than the fMRI BOLD signals ($1/\hat{t}=0.5$\,Hz). Furthermore, BOLD signals are known to be slow and delayed responses spanning over several seconds. To align the activations of the models and the fMRI BOLD signals, we thus convolve the model activations with a standard hemodynamic response function (HRF). Specifically, activations of the network $m(U) \in \bbR^{d_{\hat{t}} \times d_x}$ are normalized between $[0, 1]$. We then use nistats \cite{abraham2014machine} \texttt{compute\_regressor} function with the glover model to temporally convolve ($h\in \bbR^{d_t}$) and temporally downsample ($g: \bbR^{d_{\hat{t}}} \rightarrow \bbR^{d_t}$) each artificial neuron $j$:

\begin{equation}
    x^{(j)} = g \Big ( m^{(j)}(U) * h \Big )
\end{equation}

\subsection{Brain score}
The similarity between the preprocessed model's activations $X \in \bbR^{d_t\times d_x}$ and the BOLD response $Y\in \bbR^{d_t \times d_y}$ is assessed with a "brain score" \citep{yamins2014performance}. Specifically, we use a cross-validation within each subject (12 train-test splits across 5 sentences series) to fit an $L_2$-regularized linear model $W \in \bbR^{d_x\times d_y}$ trained to predict the BOLD time series from the model activations for each voxel independently:

\begin{equation}
    \argmin_{w}\ \sum_{i} (w^T X_{i} - y_{i})^2 + \lambda \|w\|^2
\end{equation}

with $X$ and $y$ normalized across samples to have a mean of 0 and and a variance of 1, and the training samples $i \in I$ .

Optimization is done using \texttt{RidgeCV} from scikit-learn \cite{pedregosa2011scikit}, with the penalization hyper-parameter $\lambda$ varying between 10 and $10^8$ (20 values scaled logarithmically) chosen within nested cross-validation of the training set and independently selected for each dimension.

The performance of the ridge regression is summarized with a Pearson's correlation score between the predicted and the fMRI BOLD responses on the test samples $I^*$, a.k.a "brain score":

\begin{equation}
    R = Corr(y_{I^*}, w^T \cdot X_{I^*})
\end{equation}


Note that we also compute a baseline brain-score with $X$ being the preprocessed mel filterbanks coefficients (mel) of the auditory stimuli, using librosa \cite{librosa}, 25 milliseconds windowing and strides of 10 milliseconds. These spectro-temporal decompositions were originally engineered to mimic the cochlear transformations of sounds. They thus provide a reasonable approximation of the features input to the primary auditory cortex \cite{stevens1937scale, stevens1940relation}.

\subsection{Statistics}
The reliability of brain scores is assessed with second-level analyses across subjects, by applying a Wilcoxon signed-rank test across subjects. Statistical correction for multiple comparisons is performed with Benjamini–Hochberg False Discovery Rate (FDR) across voxels \cite{benjamini2010discovering}. Figures only plot brain scores that are significant across subjects after this correction.

\section{Results}

\begin{figure*}
    \centering
    \vskip -0.2in
    \includegraphics[height=5.5in]{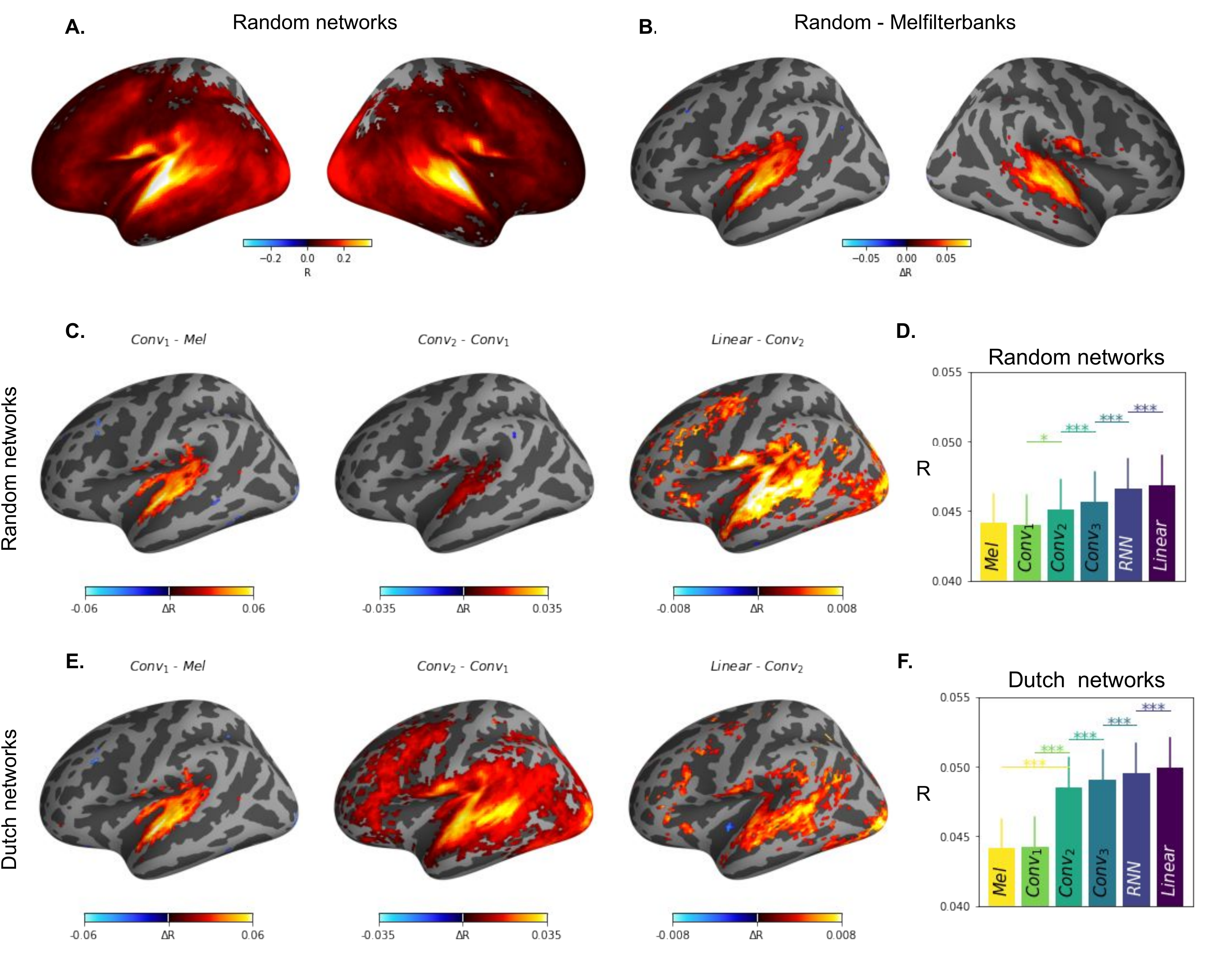}
    \caption{\textbf{A} Average (across subjects and random seeds) brain scores obtained by the random initialisation model. \textbf{B.} Average (across subjects and random seeds) gains in brain scores ($\Delta R$) between the random initialisation model and the mel filterbanks, using a concatenation of all sound features and activations. Gray zones are not statistically significant across subjects after FDR correction for multiple comparison. \textbf{C.} Average (across subjects and random seeds) gains in brain scores ($\Delta R$) obtained between four representative layers of the random initialisation models.
    \textbf{D.} Average (across voxels and random seeds) brain scores obtained for each level of concatenation for the random initialisation models. Error bars indicate SEM across subjects. Vertical lines and stars indicate significant gains across subjects. Redundant statistical comparisons are omitted for clarity purposes (e.g. $C>A$ is omitted when $C>B$ and $B>A$).
    \textbf{E. } Same as (B) for the networks trained on Dutch.
    \textbf{F.} Same as (C) for the networks trained on Dutch.}
    \label{fig:random}
    \vskip -0.2in
\end{figure*}

\subsection{Random networks reveal the inductive bias of hierarchical convolutions}

We first aim to test whether random (i.e. not trained) convolutional architectures generate representations similar to those of the brain. 
To this end, we assess, for each subject independently, whether their fMRI responses $Y$ can be predicted with a linear regression $W$ from the concatenated activations $X$ of the random intialisation model input with the same speech sounds, as quantified with a Pearson correlation, hereafter referred to as "brain scores" ($R$, Figure \ref{fig:random}.A).
Sound onset and volume are known to trigger attentional capture in large brain networks \cite{daube2019simple}. Consequently, significant brain scores result from trivial low-level features. To mitigate this issue, we compare the brain-scores obtained with all layers $L \in [1, 5]$ to those obtained with the mel filterbanks (denoted as $L_0$, Figure \ref{fig:random}.B):
\begin{equation}
        \Delta R = R(X_{L\in [0, 5]}) - R(X_{0})
\end{equation}

We perform a similar comparison for each pair of successive layers to identify the layer-specific representations that correlate with brain responses (Figure \ref{fig:random}.C):
\begin{equation}
    \Delta R_{L}=R(X_{[0, \dots, L]}) - R(X_{[0, \dots, L-1]})
\end{equation}
The results show that the first layers of the model lead to a significant gain in brain prediction around the primary and secondary auditory areas as well as in the superior temporal gyrus, whereas the deepest layers led to significant gains in a distributed set of brain areas. The average brain score obtained by the random initialisation model for each level of concatenation can be seen in Figure \ref{fig:random}.D.

Overall, these results suggest that 1) the representations generated by random convolutional networks suffice to predict a significant proportion of the brain representations of speech \emph{above-and-beyond} mel filterbanks and 2) that the hierarchy of the random neural networks automatically map onto the hierarchy of auditory processing in the cortex.

\subsection{Training on Dutch increases the brain-similarity of neural networks}

Does training make networks' representation more brain-like?
To tackle this issue, we follow the same brain-scores analysis with networks trained on a Dutch speech-to-text task (Figure \ref{fig:random}.E and \ref{fig:random}.F). Overall, Dutch models better predict brain responses than random networks (Figure \ref{fig:task}.A). Layer-wise comparison of the brain scores between the random networks and the trained networks shows that training leads to a significant gain for each layer but the first one ($p < 10^{-6}$ across all subsequent layers, Figure \ref{fig:task}.B).

\subsection{Impact of the phonetic distance between training and processed speech on brain-similarity}

Training with Dutch increases brain-similarity. Is this effect due to the generation of language-specific representations? To address this issue, we compare networks trained with either Bengali or English, while matching the architecture, training procedure and data exposure to the Dutch networks (see Section \ref{sec:task}). 

Training on English and Bengali induces a significant increase of brain-scores compared to random networks (Figure \ref{fig:task}.C). Training on English does not produce significantly higher brain scores than training on Dutch ($p=0.22$). This lack of effect is reassuring, as it confirms that Dutch models do not implicitly capture lexico-semantic features. On the other hand, the fact that English and Dutch networks achieve similar brain-scores suggests that the sublexical representations of speech in the brain are \emph{not} language specific. 

Finally, training on Bengali produces a significantly lower gain in brain scores than training on either Dutch ($p<0.01$) or English ($p<0.01$). Given that Dutch and English are closer phonetically than Dutch and Bengali \cite{schepens2020big}, these results suggests that the brain generates \textbf{phoneme-specific} representations.

\subsection{The first levels of the cortical hierarchy are most similar to sound-generic networks}

Does training on a non-speech task also increase brain-similarity?

To tackle this issue, we now focus on the networks trained on a curated acoustic scene classification task. 
On average across all voxels (and across initialization seeds), the brain scores of the acoustic scene networks are not significantly different from those of the Dutch (
$p=0.2$) and English networks(
$p=0.2$).

\begin{figure*}[h!]
    \centering
    \includegraphics[height=3in]{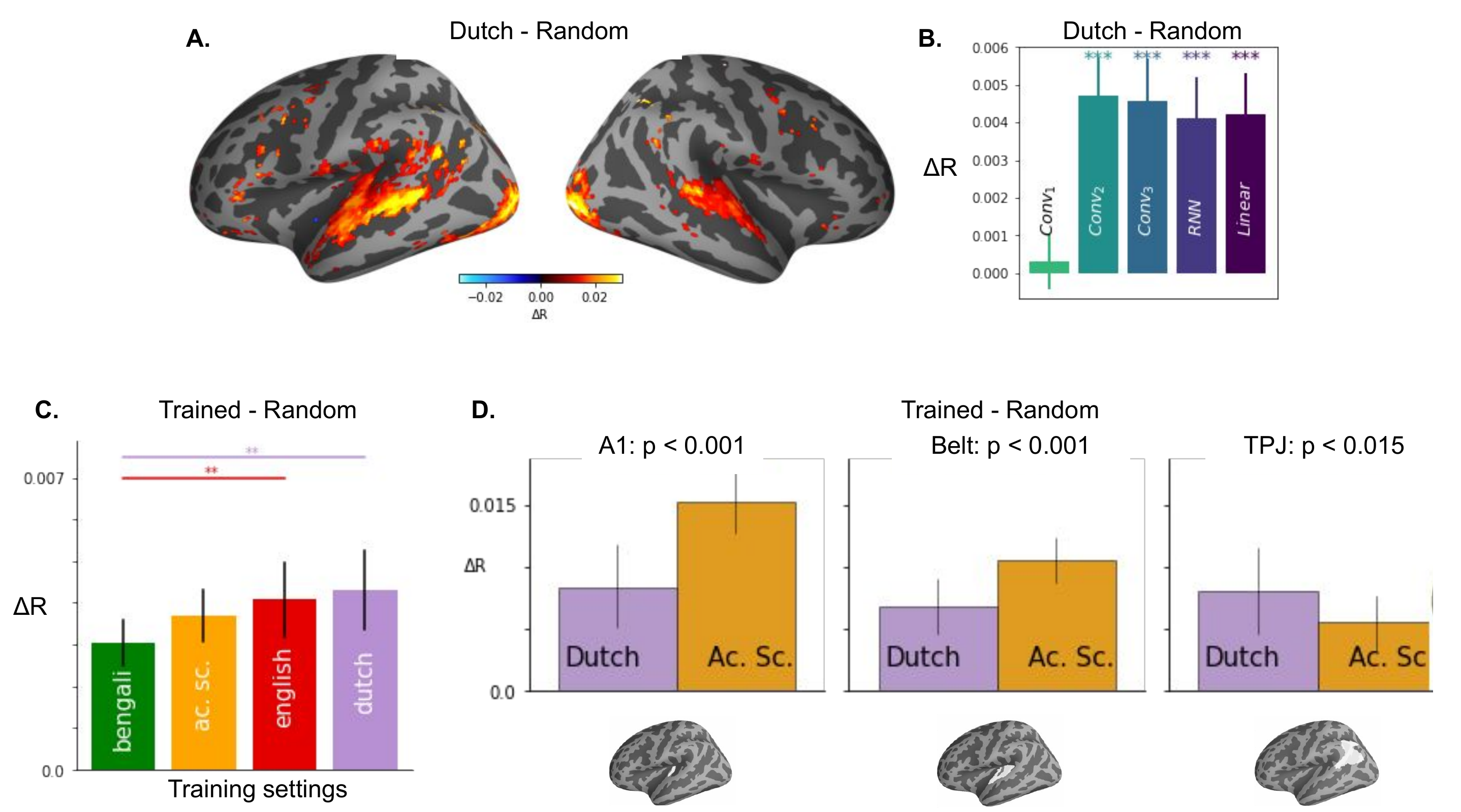}
    \caption{\textbf{A.} Average (across subjects and random seeds) gains in brain scores ($\Delta R$) between the models trained on Dutch and the random networks (untrained), using a concatenation of all sound features and activations. Gray zones are not statistically significant across subjects after FDR correction for multiple comparison. 
    \textbf{B.} Average (across voxels and random seeds) gains in brain scores between the trained model and the random model (untrained) for each layer independently. Error bars indicate SEM across subjects. Stars indicate significant gains across subjects. 
    \textbf{C.} Average improvement in brain score ($\Delta R$) obtained by the models trained --from left to right-- on Bengali, acoustic scene classification (ac. sc.), English, and Dutch, compared to their random initialisation (untrained model). Vertical lines and stars indicate significant gains across subjects. \textbf{D.} Comparison of the effect of training on Dutch and acoustic scenes for three regions of interest: the primary auditory cortex (A1), the secondary auditory cortex (Belt) and  the temporo-parietal junction (TPJ). Position of the three regions on the left hemisphere of the brain is visible under each graph.}
    \label{fig:task}
    \vskip -0.2in
\end{figure*}

However, the predictions of the brain activity in the first (A1) and secondary (Belt) auditory cortices (see Figure \ref{fig:task}.D) are significantly higher when using acoustic scene classification networks than when using English (Mean A1: 
$p<10 ^ {-9}$, Mean  Belt: 
$p< 10 ^ {-6}$) and Dutch networks (Mean A1: 
$p< 10 ^ {-7}$, Belt: 
$p< 10 ^ {-5}$). 
On the contrary, for the temporoparietal junction, where parts of the higher level processing of speech take place \citep{hickok2007cortical}, the training on Dutch leads to significantly higher gain in brain score (
Mean TPJ: $p<0.015$) than the acoustic scene classification training.

This result suggests that the brain uses sound-generic representations for the first processing stage of its hierarchy, and then builds speech-specific representations in higher-level processing stages.

\section{Discussion}

By focusing on speech representations, the present study provides four empirical contributions to the investigation of auditory representations in brains and deep learning models \cite{yamins2016using,keshishian2020estimating, berezutskaya2020brain, khosla2020cortical,kell2018task, kumar2020searching,koumura2019cascaded}.

First, deep convolutional architectures appear to already account for the hierarchy of brain responses to speech, in that their first and deepest layers linearly map onto the primary auditory and associative cortices, respectively. This result suggests that high-level auditory representations can benefit from an inductive bias imputable to a simple and biologically plausible architecture. Indeed, unlike visual conv nets, which require neurons with different visual receptive fields to share weights during training \cite{yamins2016using}, audio conv nets apply their convolution over time and frequency -- a ubiquitous operation in  sensory and associative cortices \citep{david2013integration}.

Second, training networks with a variety of auditory input appear to systematically improve brain scores. This result suggests that the brain can, in principle, use speech non-specific representations to process sounds. This finding encourages research to test the utility of pretraining speech recognition systems with non-speech data, in order to extract more (and potentially better) features.

Third, training with Dutch or on English -- i.e., a language phonetically-related to Dutch  -- achieved the highest brain scores. This results provides empirical evidence against the hypothesis that phonetic features are reducible to generic auditory representations \citep{galantucci2006motor,daube2019simple}. On the contrary, it suggests that learning speech leads the brain to generate specific phonetic representations.

Finally, training on acoustic scene classification led to higher brain scores in the primary and secondary auditory cortex than speech models. This finding supports the idea that the brain builds speech representations "on top" of generic representations, and fine-tunes of "recycles" \citep{dehaene2011unique} higher-level processing stages. 

The brain scores reported throughout the study are low. 
This limitation can be accounted for by two main factors. First, we do not report the absolute brain scores but the difference of brain scores between different layers or different models (Figure \ref{fig:random}.A for an example of full brain score, reaching up to R=0.42 in the superior temporal lobe). Second, fMRI responses are notoriously noisy, especially at the single-sample and single-voxel level. Fortunately, the large number of subjects and stimuli allows these weak effects to achieve reasonable statistical significance.



To our surprise, we observe significant brain scores in visual regions (e.g. Figure \ref{fig:task}.A). We can speculate that this phenomenon may either relate to i) feedback projection to visual areas specialized in orthographic representations \citep{dehaene2011unique} and/or ii) an auditory-dependent processing of the fixation cross \citep{petro2017contextual,majka2019unidirectional}. However, this unexpected result requires temporally-resolved brain recordings to confirm and clarify the nature of such auditory-induced visual responses.


 


Overall, our study strengthens the mutual relevance of machine learning models and neuroscience. While the present investigation remains to be generalized to more efficient audio processing models \cite{baevski2020wav2vec,collobert2016wav2letter}, it suggests, once again, that some of the computational solutions designed by machine learning researchers, may actually converge to or re-discover those implemented in the human brain \citep{marblestone2016toward, caucheteux2020language}.

\bibliography{paper}
\bibliographystyle{icml2021}

\end{document}